\documentclass[11pt,a4paper]{article}
\usepackage[hyperref]{naaclhlt2018}
\usepackage{times}
\usepackage{latexsym}
\usepackage{graphicx}
\usepackage{url}
\usepackage[noend]{algpseudocode}
\usepackage{fancyhdr,amsmath,amssymb}
\usepackage[ruled,vlined]{algorithm2e}
\aclfinalcopy % Uncomment this line for the final submission
\title{Multi-Document Summarization using Distributed Bag-of-Words Model}

\author{Kaustubh Mani \\
  IIT Kharagpur \\
  \\\And
  Ishan Verma \\
  TCS Research \\
   \\\And
  Hardik Meisheri \\
  TCS Research \\
   \\\And
  Lipika Dey \\
  TCS Research \\
  \\}

\date{}

\begin{document}
\maketitle
\begin{abstract}
  As the number of documents on the web is growing exponentially, multi-document summarization is becoming more and more important since it can provide the main ideas in a document set in short time. In this paper, we present an unsupervised centroid-based document-level reconstruction framework using distributed bag of words model. Specifically, our approach selects summary sentences in order to minimize the reconstruction error between the summary and the documents. We apply sentence selection and beam search,  to further improve the performance of our model. Experimental results on two different datasets show significant performance gains compared with the state-of-the-art baselines.
\end{abstract}

\section{Introduction}
Multi-document summarization is a process of representing a set of documents with a short piece of text by capturing the relevant information and filtering out the redundant information. Two prominent approaches to multi-document summarization are extractive and abstractive summarization. Extractive summarization systems aim to extract salient snippets, sentences or passages from documents, while abstractive summarization systems aim to concisely paraphrase the content of the documents.

In this paper, we propose a centroid-based document level reconstruction framework using distributed bag-of-words (\textit{PV-DBOW}) \cite{pv} model. Summary sentences are selected in order to minimize the reconstruction error between the summary and documents. 

% They have been applied to several natural language processing tasks like sentiment analysis (Tang et al., 2015) and topic detection (Kazuma et al., 2016)(http://www.sciencedirect.com/science/article/
% pii/S1532046416300442). We use both the paragraph vector models: "distributed memory"(DM) and "distributed bag of words"(DBOW) model and compare their results on benchmark datasets. 

In this work:
\begin{itemize}
%   \item We formulate multi-document summarization as a centroid-based document level reconstruction problem, and then use paragraph vectors to solve it.
  \item We propose the use of Distributed Bag of Words (PV-DBOW) model for multi-document summarization. 
%   \item We compare the performance of distributed memory model and distributed bag-of-words model on multi-document summarization task in an unsupervised setting.
\item Since document representation is central to the implementation of our model. We compare several document representation techniques using the document-level reconstruction framework.
  \item We conduct experiments on DUC 2006 and DUC 2007 benchmark datasets to show the improvement of our model over previous unsupervised summarization systems. 
  
%   Our method outperforms the state-of-the-art 
% unsupervised summarization system.
\end{itemize}

\section{Proposed Framework}
Several summarization methods use the bag of words (\textit{BOW}) model for sentence ranking and sentence selection  \cite{lexrank,centroid}. Bag of words model fails to encode the semantic relationship between words when comparing sentences.  Paragraph vectors \cite{pv} have been recently proposed as a method for learning fixed-length distributed representations from variable-length pieces of text.  The method has been proven to be effective for representing documents and sentences in several natural language processing tasks like sentiment classification \cite{pv}, topic detection \cite{topic} and document similarity \cite{docembed}. In this work, we use Distributed bag of words (\textit{PV-DBOW}) model to represent documents and sentences. First, we train the \textit{PV-DBOW} model to compute document vectors for all the documents in a document set then, we represent the main content of a document set by its centroid vector, which is calculated by averaging the document vectors. Summary sentences are then selected in order to minimize the reconstruction error between the documents and the summary. Sentence selection is performed to reduce the redundancy in the summary and beam search is used to further minimize the reconstruction error by exploring a large search space of candidate summaries. 

% We try to reconstruct the content of the document set by selecting sentences that minimize the reconstruction error. We use sentence selection and beam search to further improve the performance of our model.

% Paragraph Vectors was introduced by \cite{pv} to represent texts of variable lengths, such as sentences, paragraphs, and documents. They proposed two models for this task: "Distributed Memory" (PV-DM) and "Distributed Bag of Words" (PV-DBOW).In this work, we use Distributed Bag of Words (PV-DBOW) model to represent documents and sentences. 

% is similar to word2vec model proposed by Mikolov et al., 2013. In this model, every document and word is mapped to a unique vector, represented by a column in matrix D and W respectively. The document and word vectors are averaged or concatenated to predict the next word in a context. In our work, we use 

% \begin{figure}
% \includegraphics[scale=0.60]{PV-DM}
% \caption{The distributed memory model of Paragraph Vectors.}
% \end{figure}

% The model tries to minimize the log-likelihood of word $x_{i+3}$ given a document {\tt D} and words $x_i$, $x_{i+1}$, $x_{i+2}$ in a fixed context window.

\subsection{Distributed Bag of Words Model}

Distributed Bag of Words model is a simpler version of paragraph vectors, which takes the document vector \textit{D} as input and forces the model to predict words in a text window of  \textit{n} words randomly sampled from the document. 

% In other words, the model tries to maximize the log-likelihood of the context words given the document d.

% \begin{equation}
%  \mathcal{L} = \sum_{i=1}^{N} log \hspace{0.5mm} \mathbb{P}(w_i, w_{i+1}, .., w_{i+n} | d) 
%  \end{equation}

During training, document vector \textit{D} and softmax weights \textit{U} are randomly initialized and updated using stochastic gradient descent via backpropagation. At inference stage, for a new document or sentence, document vector \textit{D} is randomly initialized and updated by gradient descent while keeping the softmax weights \textit{U} fixed. Unlike distributed memory model (\textit{PV-DM}), which tries to predict the next word given a context, \textit{PV-DBOW} predicts the context directly from the document vector. This enables the model to encode higher n-gram representations, thus making it more suitable for our task of document reconstruction. In comparison to the \textit{PV-DM} version of paragraph vectors, \textit{PV-DBOW} has fewer number of parameters and thus needs less data to train.

\begin{figure}[h!]
\includegraphics[scale=0.28]{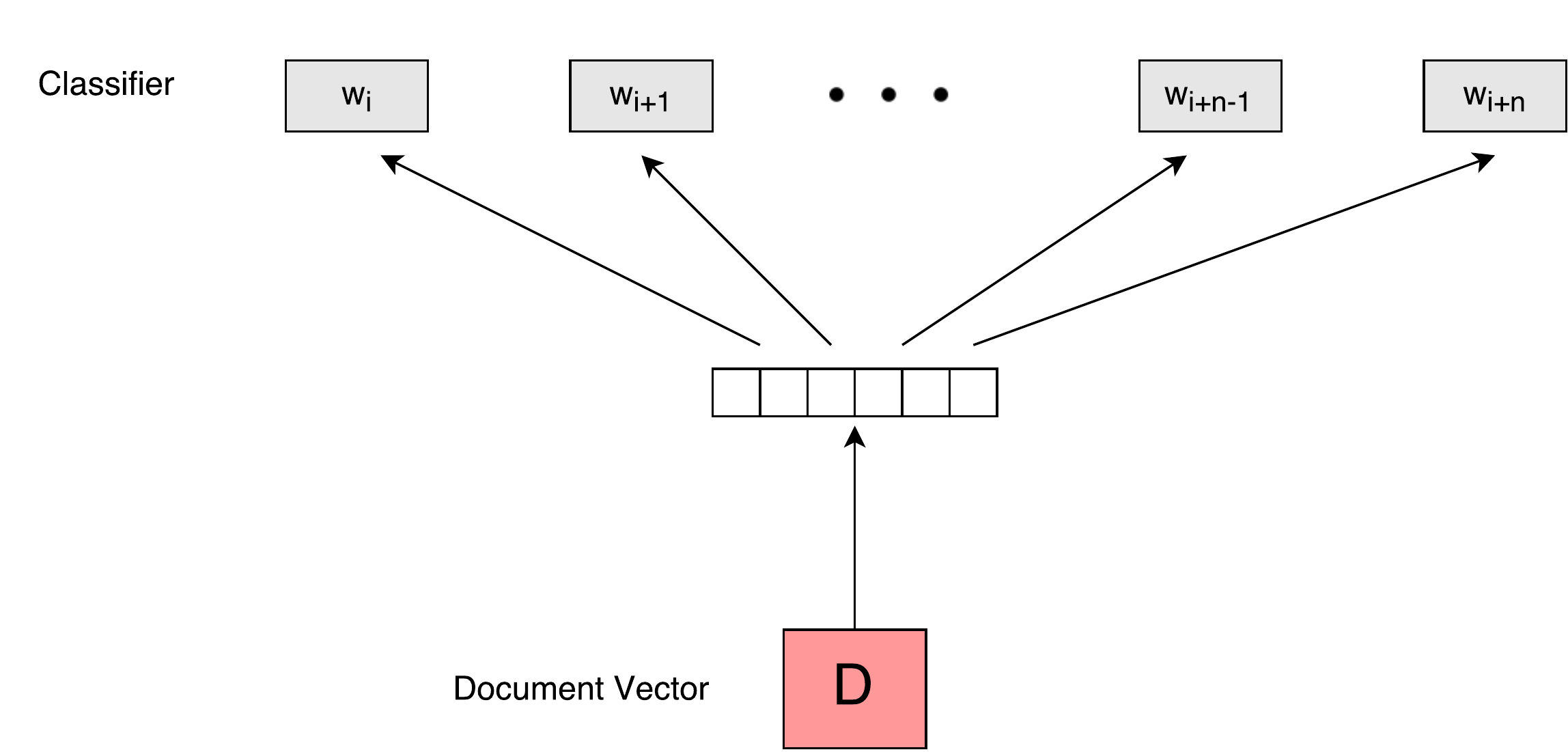}
\caption{The distributed bag of words model of Paragraph Vectors learns to predict words in a context.}
\end{figure}

\subsection{Document Reconstruction}

We treat summarization task as a multi-document reconstruction problem. We assume that a good summary is one which can reconstruct the main content of a document set. We assume that the centroid of all the documents is representative of all the meaningful content in the document set. Our assumption is inspired by \cite{centroid} where the idea was first introduced.

Given, a multi-document set D = [$d_1$, $d_2$, ... , $d_n$], centroid vector C is represented by:

% Further, to validate our assumption we randomly select three document sets and their respective model summaries from DUC 2006 dataset, compute their vector representation using PV-DBOW model and project the documents and summaries onto two-dimensional space (see Figure 3). Subsequently, we randomly select a document set and its corresponding model summaries from DUC 2007 and project centroid vector of both documents and summaries (see Figure 4). We find that the centroid of document vectors and the centroid of summary vectors are very close to each other. Hence in our framework, the main content of a document set is represented by the centroid of the document vectors.

\begin{equation}
 C = \frac{1}{n} \sum_{i=1}^{n}  DBOW(d_i) 
\end{equation}

where \textit{n} is the total number of documents in the multi-document set, and DBOW represents the Distributed Bag of Words model (\textit{PV-DBOW}). Our basic model builds the summary by iteratively selecting the sentences with the minimum reconstruction error, given by equation (2).

\begin{equation}
 ReconError = || C - \textit{DBOW}(S^*) ||_2^2
\end{equation}

where $S^*$ denotes a candidate summary. 

% \begin{figure}
% \includegraphics[scale=0.45]{DUC2007_doc_summary_centroid}
% \caption{The distributed bag of words model of Paragraph Vectors.}
% \end{figure}

\subsection{Sentence Selection}
 Given a document set, we create a candidate set of sentences S=[$s_1$, $s_2$, .., $s_N$], which contains all the sentences in the document set. Sentence vectors for all the sentences in the candidate set are computed using the trained \textit{PV-DBOW} model. 
The sentences in the candidate set are sorted according to their reconstruction error given by (2). Reconstruction error is minimized by iteratively selecting sentences from the candidate set into the summary set until the summary length exceeds a max limit given by \textit{K}. At each iteration, we calculate the cosine similarity between candidate sentence vector and the sentence vectors of the sentences which are already present in the summary set. The sentence having cosine similarity greater than a threshold $\theta$ are not selected in the summary set. 
\begin{equation}
 sim(s_i, s_j) = \frac{s_i \cdot{} s_j^T}{||s_i|| \cdot ||s_j|| }
\end{equation}

% In Algorithm 1, S, ReconError, $\theta$, K, DBOW represent the candidate set, reconstruction error, similarity threshold, summary length and distributed bag of words model respectively. 
% \makeatletter
% % \def\BState{\State\hskip-\ALG@thistlm}
% \makeatother

\begin{algorithm}[h]
\KwIn{S, ReconError, $\theta$, K, DBOW}
\KwOut{Summary}
\hspace{3.5mm} Summary $\leftarrow$ $\emptyset$\\
\hspace{3.5mm} S $\leftarrow$ SORT(S, ReconError) \\
\hspace{3.5mm} {\bf for} sentence $s_c$ in S {\bf do}  \\
\hspace{9mm}   {\bf if} len(Summary) $\textgreater$ K {\bf then} \\
\hspace{14.5mm}  {\bf return} Summary \\
\hspace{9mm}   $sv_c$ $\leftarrow$ DBOW($s_c$) \\
\hspace{9mm}   select $\leftarrow$ True  \\
\hspace{9mm}   {\bf for} $s_s$ in Summary {\bf do} \\
\hspace{14.5mm}  $sv_s$ $\leftarrow$ DBOW($s_s$) \\
\hspace{14.5mm}  {\bf if} sim($sv_c$, $sv_s$) $\textgreater$ $\theta$ {\bf then} \\
\hspace{20mm}   select $\leftarrow$ False \\
\hspace{9mm} {\bf if} select {\bf then} \\
\hspace{14.5mm} Summary $\leftarrow$ Summary $\cup$ $s_c$

\caption{{ Sentence Selection} \label{Sentence Selection}}

\end{algorithm}

\subsection{Beam Search}

Beam search is a heuristic state space search algorithm, which is basically a modification of breadth first search. The algorithm loops over the entire candidate set \textit{S} and selects sentences until the summary length exceeds the max length limit given by \textit{K}.
At each iteration, sentences in candidate set are added to the summaries present in the summary set, the vectors for each summary is computed using trained \textit{PV-DBOW} model and reconstruction error is calculated. The summaries present in the summary set are sorted according to their reconstruction error and only top \textit{k} summaries are retained in the summary set for the next iteration. \textit{k} is often referred as beam width. After the algorithm terminates summary set containing \textit{k} summaries is returned. Out of these, we consider the summary with the minimum reconstruction error as the output of beam search algorithm.

\section{Experiments}

We conducted experiments with two standard summarization benchmark datasets DUC 2006 and DUC 2007 provided by NIST \footnote{http://www/nist.gov/index.html} for evaluation. DUC 2006 and DUC 2007 contain 50 and 45 document sets respectively. Each document set consists of 25 news articles and 4 human-written summaries as ground truth. The summary length is limited to 250 words (whitespace delimited).

\subsection{Implementation}

Neural Network based models are difficult to train on small datasets. For this purpose we train our model on a combined corpus of Thomson Reuters Text Research Collection (TRC2) in Reuters Corpora \cite{trc2} and CNN/Dailymail dataset first, and then fine tune on DUC 2006 and DUC 2007 datasets. We use gensim\footnote{https:radimrehurek.com/gensim/index.html} library in python to train our \textit{PV-DBOW} model.The hyper-parameters of the model are selected through parameter tuning on DUC 2005 dataset using grid search in the following setting: paragraph vector size in [100, 200, 300, 400, 500], window size in [8, 10], similarity threshold in [0.5, 1] with a step of 0.05, and beam width in [5, 40] in steps of 5.
\begin{table}[ht]
\caption{\% Average F-measure on DUC 2007} % title of Table
\centering 
\resizebox{\columnwidth}{!}{%
\begin{tabular}{|l c c r|}
	\hline
	{\bf Model} & {\bf Rouge-1} & {\bf Rouge-2} & {\bf Rouge-SU4}\\
    \hline 
    CBOW  & 38.649 & 7.942 & 13.584 \\
%     \hline 
    PV-DM &  39.826 & 8.514 & 13.875 \\
%     \hline
    PV-DBOW & \textbf{42.679} & \textbf{10.916} & \textbf{16.320} \\
    \hline
\end{tabular}%
\label{table:nonlin} % is used to refer this table in the text
}
\end{table}
\subsection{Evaluation Metric}

We run the ROUGE (Recall-Oriented Understudy for Gisting Evaluation) metrics \cite{rouge} which ROUGE measures summary quality by counting overlapping units such as n-grams word sequences and word pairs between the generated summary(produced by algorithms) and the model summary (human labeled). We choose ROUGE-N and ROUGE-SU4 in our experiments. Formally, ROUGE-N is an n-gram recall and ROUGE-SU4 is an unigram plus skip-bigram match with maximum skip distance of 4 between between a system generated summary and a set of model summaries.

\section{Compared Methods}

As our framework is unsupervised, we compare our model with state-of-the-art unsupervised summarization systems. Document reconstruction based methods like \textit{SpOpt} \cite{spopt}, \textit{DocRebuild} \cite{docrebuild} and \textit{DSDR} \cite{dsdr} are the direct baselines for comparison. \textit{SpOpt} uses a sparse representation model which selects sentences and does sentence compression simultaneously. \textit{DocRebuild} uses distributed memory (\textit{PV-DM}) model to represent documents and selects sentences using a document level reconstruction framework. \textit{DSDR} selects sentences from the candidate set by linearly reconstructing all the sentences in the document set, and minimizes the reconstruction error using sparse coding. We also show two weaker baselines \textit{Random} and \textit{Lead} \cite{lead}. \textit{Random} does a random selection of sentences for each document set. \textit{Lead} sorts the documents in a document set chronologically and  selects the leading sentences from each documents one by one. We use \textit{PV-DBOW} to denote our basic model, \textit{PV-DBOW + SS} to denote our model with sentence selection, \textit{PV-DBOW+BS} to denote our model with beam search.

We also compare \textit{PV-DBOW} model with other document representation techniques like \textit{CBOW} and \textit{PV-DM} using the same document reconstruction framework.

% \footnote{DocRebuild also uses other summarization systems like SpOpt, DSDR to improve the performance of their model. For fair comparision, we have implemented only the basic version of their model.}

\begin{figure}
\includegraphics[width = 0.9\linewidth]{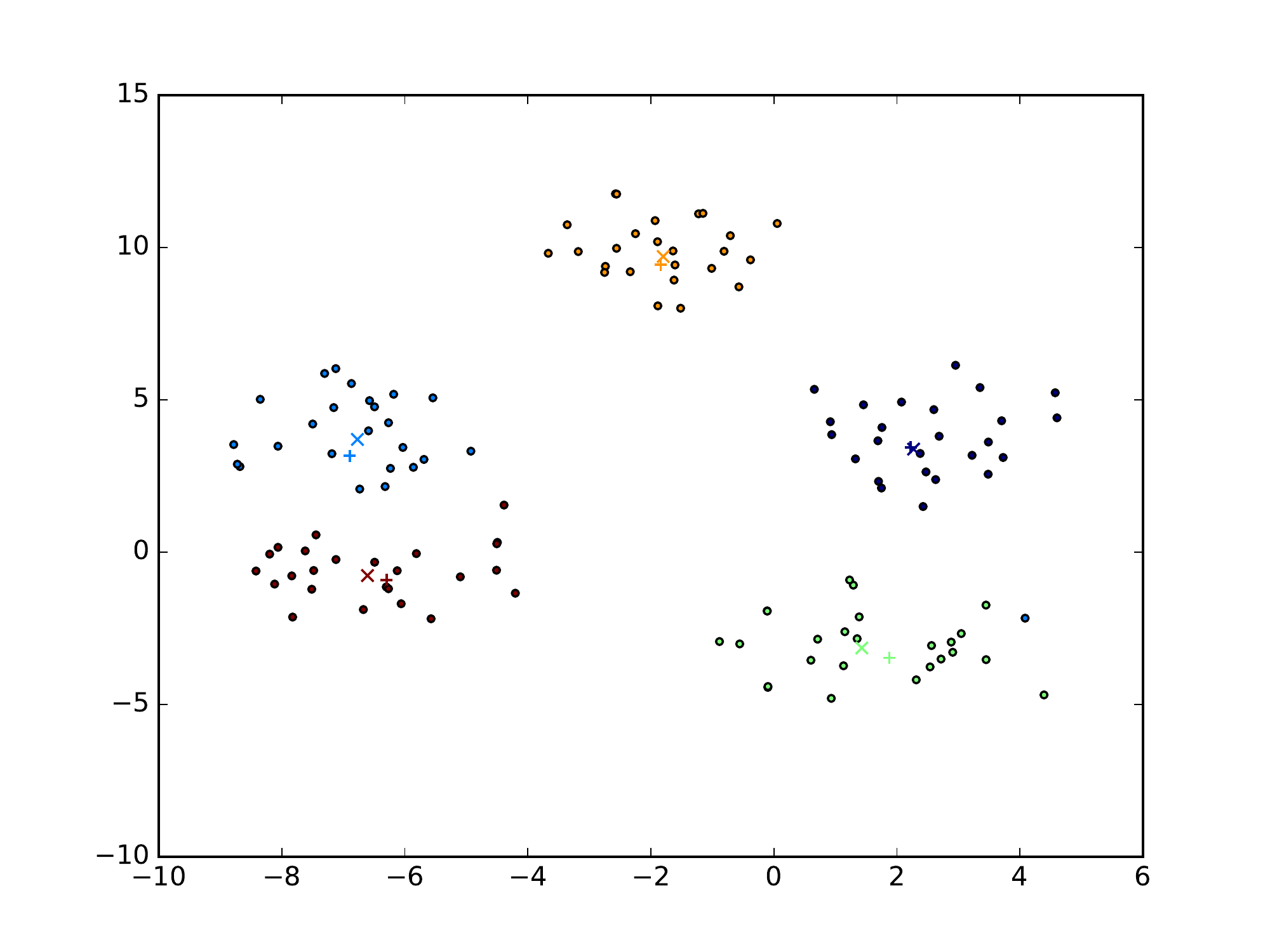}
\caption{Visualization of summaries created by our model (x) and the Centroid of reference summaries (+) for five document sets in DUC 2006.}
\end{figure}
% \begin{itemize}
%   \item \textbf{Random}: randomly selects sentences for each document set.
%   \item \textbf{Lead} (Wasson 1998): chronologically sorts the documents in a document set and selects the leading sentences from each documents one by one.
% %   \item \textbf{LexRank} (Erkan and Radev, 2004): uses the TextRank algorithm (Mihalcea and Tarau, 2004) to establish a ranking between sentences.
%   \item \textbf{DSDR} (He et al. 2012): selects sentences from the candidate set by linearly reconstructing all the sentences in a document set, and minimizing the reconstruction error using sparse coding.
  
%   \item \textbf{SpOpt} (Yao et al., 2015): uses a sparse representation model which selects sentences and does sentence compression simultaneously.
%   \item \textbf{DocRebuild} (Shulei et al. 2016): uses distributed memory model (PV-DM) to represent documents and selects sentences using centroid-based document level reconstruction.
% \end{itemize}

\section{Results and Discussion}
\label{sec:blind}

The results for all the experiments performed on DUC 2006 and DUC 2007 datasets are shown in Table 1 and Table 2 respectively. 
As shown in the table, \textit{Random} and \textit{Lead} give the poorest performance. \textit{DSDR} improves the performance by introducing a data reconstruction based system. \textit{DocRebuild} performs better by using a document level reconstruction framework. \textit{SpOpt} improves the performance even further by doing sentence compression and putting the diversity constraint. Our basic model outperforms all the baselines and \textit{PV-DBOW} with beam search achieves the best performance. It can be seen that the improvement in Rouge-2 and Rouge-SU4 scores is more significant in comparision to Rouge-1 scores. Higher Rouge-2 and Rouge-SU4 scores suggest that our model is more capable at handling n-grams than words.
% \textit{PV-DBOW} model, since the training procedure for \textit{PV-DBOW} model is not on word level unlike  models. 

% PV-DBOW model learns to predict the context and thus capture higher level n-gram representations for document and sentences. 

% \footnote{https://code.google.com/p/word2vec}

To show the effectiveness of our model, we randomly pick 5 document sets from DUC 2006 dataset and compute the vectors for our model generated summaries, and reference summaries. For each document set we plot the documents along with the system generated summary and the centroid of the 4 reference summaries. In Figure 2, each color corresponds to a document set, system generated summaries are denoted by (x), and centroids of reference summaries are denoted by (+). It can be seen from the figure that our system generates summaries are very close to the centroid of the reference summaries for each document set. 

Experimental results (Table 1) also show that \textit{PV-DBOW} is a better model for representing documents and sentences in comparision to \textit{PV-DM} \cite{docrebuild} and \textit{CBOW} at the task of document reconstruction based multi-document summarization. 
\begin{table}[ht]
\caption{\% Average F-measure on DUC 2006} % title of Table
\centering 
\resizebox{\columnwidth}{!}{%
\begin{tabular}{|l|c|c|r|}
	\hline
	{\bf Model} & {\bf Rouge-1} & {\bf Rouge-2} & {\bf Rouge-SU4}\\
	\hline
    
    Random & 33.879 & 5.184 & 10.092\\
	\hline
    Lead & 34.892 & 6.539 & 11.148\\
	\hline
%     LexRank &     &    & \\
%     \hline
    DSDR & 35.484    &  6.142  & 11.834 \\
    \hline 
    DocRebuild & \textbf{42.193} & 9.314 & \textbf{15.177}\\
    \hline 
    SpOpt &  40.418   &  8.388  & 14.232 \\
    \hline 

%     PV-DM + SS &  &  & \\
%     \hline
%     PV-DM + BS & & &\\
%     \hline
    PV-DBOW & 41.282 & 9.269 & 15.040 \\
    \hline 
    PV-DBOW + SS & 41.400 & 9.299 & 14.895 \\
    \hline 
    PV-DBOW + BS & 41.421 & \textbf{9.418} & 14.976 \\
    \hline
\end{tabular}%
\label{table:nonlin} % is used to refer this table in the text
}
\end{table}

\begin{table}[ht]
\caption{\% Average F-measure on DUC 2007} % title of Table
\centering 
\resizebox{\columnwidth}{!}{%
\begin{tabular}{|l|c|c|r|}
	\hline
	{\bf Model} & {\bf Rouge-1} & {\bf Rouge-2} & {\bf Rouge-SU4}\\
	\hline
    
    Random & 34.279 & 5.822 & 10.092\\
	\hline
    Lead & 36.367 & 8.361 & 12.973\\
	\hline
%     LexRank &     &    & \\
%     \hline
    DSDR & 37.351    &  7.892  & 12.936 \\
    \hline 
     DocRebuild & \textbf{43.426} & 10.500 & 16.246 \\
    \hline 
    SpOpt &  41.674   &  9.905  & 15.665 \\
    \hline 
%     PV-DM + SS &  &  & \\
%     \hline
%     PV-DM + BS & & &\\
%     \hline
    PV-DBOW & 42.679 & 10.916 & 16.320 \\
    \hline 
    PV-DBOW + SS & 42.617 & 11.124 & 16.462 \\
    \hline 
    PV-DBOW + BS & 42.723 & \textbf{11.231} & \textbf{16.508} \\
    \hline
\end{tabular}%
\label{table:nonlin} % is used to refer this table in the text
}
\end{table}

\section{Related Work}

Our model is closely related to data reconstruction based summarization which was first proposed by  \cite{dsdr}. Since then, several other data reconstruction \cite{spopt,docrebuild} based approaches has been proposed. \cite{mdsparse} proposed a two-level sparse representation model to reconstruct the sentences in the document set subject to a diversity constraint. \cite{nmf} proposed  a model based on Nonnegative matrix factorization (\textit{NMF}) to group the sentences into clusters. Recently, several neural network based models have been proposed for both extractive \cite{attsum,summarunner} and abstractive summarization \cite{namas,abs}

% Most of the extractive summarization systems use a ranking model to select sentences from a candidate set. (Lin and Hovy, 2002; Radev et al. 2004) used sentence clustering, (Erkan and Radev, 2004; Mihalcea and Tarau, 2005) used PageRank and (Harabagiu and Lacatusu, 2005; Wang et al., 2008) used topic modelling to rank the sentences. The problem with these methods is that top ranked sentences often share a lot of redundant information. Reducing the redundancy by selecting sentences which have both good coverage and minimum redundancy is non-trivial task.

% Data reconstruction based approaches try to overcome the problem of redundancy (He et al. 2012; Liu et al., 2015; Yao et al.,2015) by selecting a sparse subset of sentences which can linearly reconstruct all the sentences in the document set. Although the mathematical idea behind these methods is intuitive, sentence level reconstruction loses the global context of the document.

\section{Conclusion and Future Work}

In this paper, we present a document level reconstruction framework based on distributed bag of words model (\textit{PV-DBOW}). The main content of the document set is represented by a centroid vector which is computed using \textit{PV-DBOW} model, and summary sentences are selected in order to minimize the reconstruction error. We do sentence selection and beam search to further improve the performance of our model. Our model outperforms the state-of-the-art unsupervised systems and shows significant improvements over Rouge-2 and Rouge-SU4 scores. Since paragraph vectors can be used to model variable-length texts, our model can be extended to a phrase level extraction based summarization system.

\bibliography{bib}
\bibliographystyle{acl_natbib}

\appendix

\end{document}